\def\year{2022}\relax
\documentclass[letterpaper]{article} 
\usepackage{aaai22}  
\usepackage{times}  
\usepackage{helvet}  
\usepackage{courier}  
\usepackage[hyphens]{url}  
\usepackage{graphicx} 
\usepackage{natbib}  
\usepackage{caption} 
\DeclareCaptionStyle{ruled}{labelfont=normalfont,labelsep=colon,strut=off} 
\usepackage{algorithm}
\usepackage{algorithmic}

\usepackage{booktabs}
\usepackage{pifont}
\usepackage{xcolor,colortbl}
\usepackage{float}
\definecolor{darkergreen}{RGB}{21, 152, 56}
\definecolor{red2}{RGB}{252, 54, 65}

\definecolor{citecolor}{RGB}{34, 149, 34}
\usepackage[pagebackref=true,breaklinks=true,letterpaper=true,colorlinks,citecolor=citecolor,bookmarks=false]{hyperref}

\usepackage{amsmath,amssymb}
\usepackage{amsfonts}       
\usepackage{nicefrac}       
\usepackage{microtype}      
\usepackage{xcolor}         
\usepackage{wrapfig}
\usepackage{caption}
\usepackage{pifont}
\usepackage{adjustbox}
\setcitestyle{square,comma}
\usepackage{listings}
\usepackage{caption}
\usepackage{array}
\usepackage{subfig}

\usepackage{xcolor}
\usepackage{colortbl}
\definecolor{mygraylite}{gray}{.94}
\definecolor{mygray}{gray}{.89}
\definecolor{darkergreen}{RGB}{21, 152, 56}

\newcommand{\randinit}{\tablestyle{1pt}{1} \begin{tabular}{z{21}y{26}} \multicolumn{2}{c}{\demph{random init.}} \end{tabular}}

\newcolumntype{x}[1]{>{\centering\arraybackslash}p{#1pt}}
\newcolumntype{y}[1]{>{\raggedright\arraybackslash}p{#1pt}}
\newcolumntype{z}[1]{>{\raggedleft\arraybackslash}p{#1pt}}
\newlength\savewidth\newcommand\shline{\noalign{\global\savewidth\arrayrulewidth
  \global\arrayrulewidth 1pt}\hline\noalign{\global\arrayrulewidth\savewidth}}
\newcommand{\tablestyle}[2]{\setlength{\tabcolsep}{#1}\renewcommand{\arraystretch}{#2}\centering\footnotesize}

\usepackage{enumitem}
\usepackage{bm}

\usepackage{newfloat}
\usepackage{listings}
\floatstyle{ruled}
\newfloat{listing}{tb}{lst}{}
\floatname{listing}{Listing}
\title{Un-Mix: Rethinking Image Mixtures for Unsupervised Visual \\Representation Learning}
\author{
    Zhiqiang Shen\textsuperscript{\rm 1,3}, Zechun Liu\textsuperscript{\rm 1}, Zhuang Liu\textsuperscript{\rm 2}, Marios Savvides\textsuperscript{\rm 1}, Trevor Darrell\textsuperscript{\rm 2} and Eric Xing\textsuperscript{\rm 1,3}\\
}
\affiliations{
    \textsuperscript{\rm 1} Carnegie Mellon University
    \textsuperscript{\rm 2} University of California, Berkeley\\
    \textsuperscript{\rm 3} Mohamed bin Zayed University of Artificial Intelligence\\
    \{zhiqians,zechunl,marioss\}@andrew.cmu.edu, zhuangl@berkeley.edu, trevor@eecs.berkeley.edu, epxing@cs.cmu.edu
}

\usepackage{bibentry}

\begin{document}

\maketitle

\begin{abstract}
The recently advanced unsupervised learning approaches use the siamese-like framework to compare two ``views'' from the same image for learning representations. Making the two views distinctive is a core to guarantee that unsupervised methods can learn meaningful information. However, such frameworks are sometimes fragile on overfitting if the augmentations used for generating two views are not strong enough, causing the over-confident issue on the training data. This drawback hinders the model from learning subtle variance and fine-grained information. To address this, in this work we aim to involve the {\em soft distance concept} on label space in the contrastive-based unsupervised learning task and let the model be aware of the soft degree of similarity between positive or negative pairs through mixing the input data space, to further work collaboratively for the input and loss spaces. Despite its conceptual simplicity, we show empirically that with the solution -- {\bf \underline {Un}}supervised image {\bf \underline {mix}}tures (Un-Mix), we can learn subtler, more robust and generalized representations from the transformed input and corresponding new label space. Extensive experiments are conducted on CIFAR-10, CIFAR-100, STL-10, Tiny ImageNet and standard ImageNet with popular unsupervised methods SimCLR, BYOL, MoCo V1\&V2, SwAV, {\em etc}. Our proposed image mixture and label assignment strategy can obtain consistent improvement by 1$\sim$3\% following exactly the same hyperparameters and training procedures of the base methods. Code is publicly available at \url{https://github.com/szq0214/Un-Mix}.
\end{abstract}

\section{1. Introduction}
\label{sec:intro}

  Unsupervised visual representation learning has attracted increasing attention~\cite{noroozi2016unsupervised,zhang2016colorful,oord2018representation,hjelm2018learning,gidaris2018unsupervised,he2019momentum,chen2020simple,kim2020mixco,grill2020bootstrap,caron2020unsupervised,kalantidis2020hard} due to its enormous potential of being free from human-annotated supervision, {\em i.e.}, its extraordinary capability of leveraging the boundless unlabeled data. Previous studies in this field address this problem mainly in two directions: one is realized via a heuristic {\em pretext} task design that applies a transformation to the input image, such as colorization~\cite{zhang2016colorful}, rotation~\cite{gidaris2018unsupervised}, jigsaw~\cite{noroozi2016unsupervised}, {\em etc.}, and the corresponding labels are derived from the properties of the transformation on the unlabeled data. 
Another direction is contrastive learning based approaches~\cite{he2019momentum,chen2020simple} in the latent feature space, such as maximizing mutual information between different views~\cite{bachman2019learning,tian2019contrastive}, momentum contrast learning~\cite{he2019momentum,MoCov2} with instance discrimination task~\cite{wu2018unsupervised,ye2019unsupervised}, larger batch sizes and nonlinear transformation~\cite{chen2020simple}, symmetrized distance loss without negative pairs~\cite{grill2020bootstrap}, contrasting cluster assignment~\cite{caron2020unsupervised}. SimSiam~\cite{chen2020exploring} further found stop-gradient is critical to prevent from collapsing. These methods have  shown great promise on this task, achieving state-of-the-art accuracy. However, these methods focus more on designing the training frameworks and loss formulations, ignoring crucial correlations between the input and loss spaces to enable fine-grained degrees of soft similarities between positive or negative pairs in the siamese-like unsupervised frameworks.

The motivation of our work stems from some simple observations of {\em label smoothing} in supervised learning~\cite{szegedy2016rethinking}. Interestingly, it can be observed from visualizations of previous literature~\cite{muller2019does,shen2020label} that {\em label smoothing} tends to force the output prediction of networks being less confident ({\em i.e.}, lower maximum probability of predictions), but the model representation and overall accuracy still increase significantly. The explanation for this seemingly contradictory phenomenon is that with {\em label smoothing}, the learner is encouraged to treat each incorrect instance/class as equally probable. Thus, more patterns are enforced to be explored in latent representations, enabling less variation across predicted instances and/or across semantically similar samples. This further prevents the network from overfitting to the training data. Otherwise, the network will be biased to produce over-confident predictions when evaluated on slightly different test samples. Consider that contrastive learning with InfoNCE loss is essentially classifying positive congruent and negative incongruent pairs with cross-entropy loss, such an observation reveals that a typical contrastive-based method can also encounter the over-confidence problem as in supervised learning.

\vspace{-0.05in}
\noindent{\textbf{Perspective of input and label spaces on un-/self-supervised learning.}} Contrastive learning methods adopt instance classification pretext, the features from different transformations (data augmentation) of the same images are compared directly to each other. The label of each image pair is binary (positive or negative) or continuous distance metrics. Augmentation is used as a transformation to make the distance of the same image to be larger. Different from data augmentation that enlarges the dissimilar distance but the label for calculating loss is still unchanged, our proposed mixtures will manipulate the semantic distance between two images, while adjusting the label for unsupervised loss accordingly. In other words, {\em data augmentation only changes the distance of input space}, {\em i.e.}, heavier data augmentation makes two images look more different, {\em but remains unchanged in label space} in training. However, mixture will modulate both input and label spaces simultaneously and the degree of change is controllable, which can further help capture the fine-grained representations from the unlabeled images and force models to learn more precise and smoother decision boundaries on the latent features. As a result, neural networks trained with new spaces learn flattering class-agnostic representations, that is, with fewer directions of variance on semantically similar classes. The mechanism of image mixtures in unsupervised learning is generally different from the data augmentation. 
Whereas, from the perspective of enlarging the training data space, mixtures can be considered as a broader concept of augmentation scheme in unsupervised learning.

\begin{figure}[t]
  \centering 
    \centering 
     \includegraphics[width=0.43\textwidth]{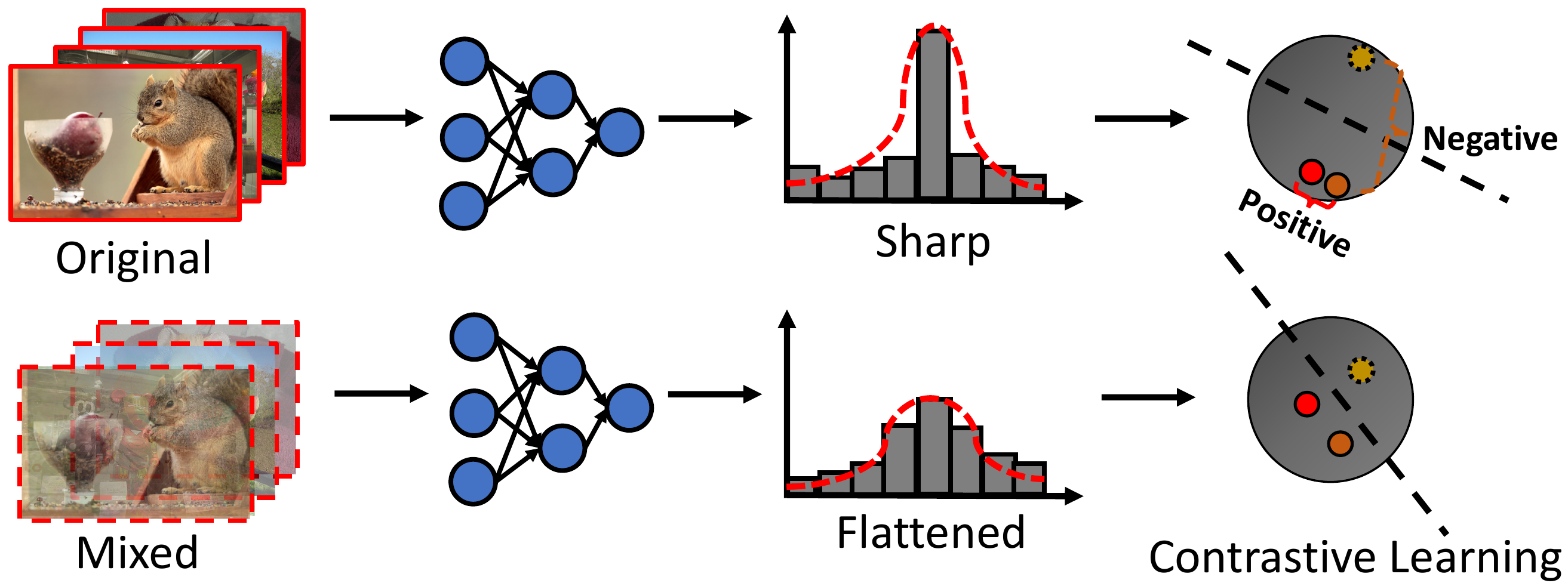}
    \label{fig:side:caption} 
      \vspace{-0.1in}
     \caption{\small{Illustration of the motivation in this work. We take the contrastive-based unsupervised learning approaches as an example. Contrastive learning measures the similarity of sample pairs in the latent representation space. With {\em flattened prediction/label}, the model is encouraged to treat semantically similar/dissimilar instances as equally probable, which will smooth decision boundaries and prevent the learner from becoming over-confident.}}
  \vspace{-0.1in}
\end{figure}

We verify our method on five recently proposed unsupervised learning methods:  SimCLR~\cite{chen2020simple}, MoCo V1\&V2~\cite{he2019momentum,MoCov2}, BYOL~\cite{grill2020bootstrap}, SwAV~\cite{caron2020unsupervised} and Whitening~\cite{ermolov2020whitening} as our baseline approaches. We conduct extensive experiments on CIFAR-10, CIFAR-100, STL-10, Tiny/standard ImageNet classification, as well as downstream object detection task on PASCAL VOC and COCO to demonstrate the effectiveness of our proposed approach. We observe that our mixture learned representations are extraordinarily effective for the downstream  detection task which empirically proves that our method can improve the model's generalizability. For instance, our 200-epoch trained model outperforms the baseline MoCo V2 by 0.6\% ($\text{AP}_{50}$), and is even better than the MoCo V2 800-epoch model.

Our contributions are summarized as follows:
\vspace{-0.04in}
\begin{itemize}[leftmargin=0.2in]
	\addtolength{\itemsep}{-0.01in}
	\item We provide empirical analysis to reveal that {\em mixing input images} and {\em smoothing labels} could improve performance favorably for a variety of unsupervised learning methods. We applied two simple image mixture methods based on previous literature~\cite{zhang2018mixup,yun2019cutmix} to encourage neural networks to predict less confidently. 
	\item We show that input and label spaces matter. We provide empirical evidence on how flattening happens under ideal conditions of latent space, validate it empirically on practical situations of contrastive learning, connect it to previous works on analyzing the discipline inside the unsupervised learning behavior. We explain the difficulties raised with original image space when visualizing distributions of predictions. Thus, we conclude that {\em good input and label spaces} are crucial for unsupervised optimization.
	\item Our proposed method is simple, flexible and universal. It can be utilized in nearly all mainstream unsupervised representation learning methods and only requires {\em a few lines of PyTorch codes} to incorporate in an existing framework. We demonstrate with a variety of base approaches and datasets, including SimCLR, BYOL, MoCo V1\&V2, SwAV, {\em etc.}, on CIFAR-10, CIFAR-100, STL-10, Tiny ImageNet and ImageNet. Our method obtains consistent accuracy improvement by 1$\sim$3\% across them. 
\end{itemize}

\section{2. Related Work}

\noindent{\textbf{(i) Un/Self-supervised Visual Feature Learning.}}
Unsupervised learning aims to exploit the internal distributions of data and learn a representation without human-annotated labels. To achieve this purpose, early works mainly focused on reconstructing images from a latent representation, such as autoencoders~\cite{vincent2008extracting,vincent2010stacked,masci2011stacked}, sparse coding~\cite{olshausen1996emergence}, adversarial learning~\cite{goodfellow2014generative,donahue2016adversarial,donahue2019large}. After that, more and more studies tried to design handcrafted pretext tasks such as image colorization~\cite{zhang2016colorful,zhang2017split}, solving jigsaw puzzles~\cite{noroozi2016unsupervised}, counting visual primitives~\cite{noroozi2017representation}, rotation prediction~\cite{gidaris2018unsupervised}. Recently, contrastive-based visual representation learning~\cite{hadsell2006dimensionality} has attracted much attention and achieved promising results. For example, Oord et al.~\cite{oord2018representation} proposed to use autoregressive models to predict the future samples in latent space with probabilistic contrastive loss.  
Hjelm et al.~\cite{hjelm2018learning} proposed to maximize mutual information from the encoder between inputs and outputs of a deep network. Bachman et al.~\cite{bachman2019learning} further extended this idea to multiple views of a shared context. 
Moreover, He et al.~\cite{he2019momentum} proposed to adopt momentum contrast to update the models and Misra\&Maaten~\cite{misra2019self} developed the pretext-invariant representation learning strategy that learns invariant representations from the pre-designed pretext tasks. The clustering-based methods~\cite{caron2018deep,caron2020unsupervised} are also a family for the unsupervised visual feature learning. 
\noindent{\textbf{(ii) Smoothing Label/Prediction in Supervised Learning.}}
Explicit label smoothing has been adopted successfully to improve the performance of deep neural models across a wide range of tasks, including image classification~\cite{szegedy2016rethinking}, object detection~\cite{krothapalli2020adaptive}, machine translation~\cite{vaswani2017attention}, and speech recognition~\cite{chorowski2016towards}. Moreover, motivated by mixup, Verma et al.~\cite{verma2019manifold} proposed to implicitly interpolate hidden states as a regularizer that encourages neural networks to predict less confidently (softer prediction) on interpolations of hidden representations. They found that neural networks trained with this kind of operation can learn flatter class representations that possess better generalization, as well as better robustness to novel deformations and even adversarial examples in testing data. Some recent work~\cite{muller2019does,shen2020label} further demonstrated that label smoothing implicitly calibrates the prediction of learned networks, so that the confidence of their outputs are more aligned with the true labels of the trained dataset. However, all of these studies lie in supervised learning.
\textbf{(iii) Differences to {\em i}-Mix}~\cite{lee2021imix} and {\bf MixCo}~\cite{kim2020mixco}. These two concurrent works also employ the idea of image mixtures on unsupervised learning but the similarity to our Un-Mix is more in the spirit than the concrete solution. The mechanism and implementation are generally different. We achieve the mixture operation by using a {self}-mixture strategy within a {\em mini}-batch of samples during training, which is simpler and more manageable for incorporating the proposed method into the existing unsupervised frameworks for mixture purpose.

\section{3. Our Approach}

In this section, we begin by presenting different paradigms using mixtures in the unsupervised learning framework, including mixing both two branches and a single one. Then, we discuss image mixture strategies and the circumstances that contain a memory bank or not. Lastly, we elaborate the loss functions for our approach and provide analysis for explaining the information gain of our proposed method.

\noindent{\textbf{Conventional siamese-like framework for unsupervised learning}.}
Given an image $I$, we first augment it to two transformed views $I_A$ and $\hat I_A$ by applying a pre-defined random transformation. Then, we feed into a two-branch framework with a projection head to produce latent representations. Finally, we define metric loss, such as InfoNCE, distance losses for optimization, as shown in Fig.~\ref{fig:paradigms} (1).

\vspace{-0.1in}
\subsection{3.1. Paradigms of Mixtures}
\begin{figure}[t]
  \centering
  \includegraphics[width=0.42\textwidth]{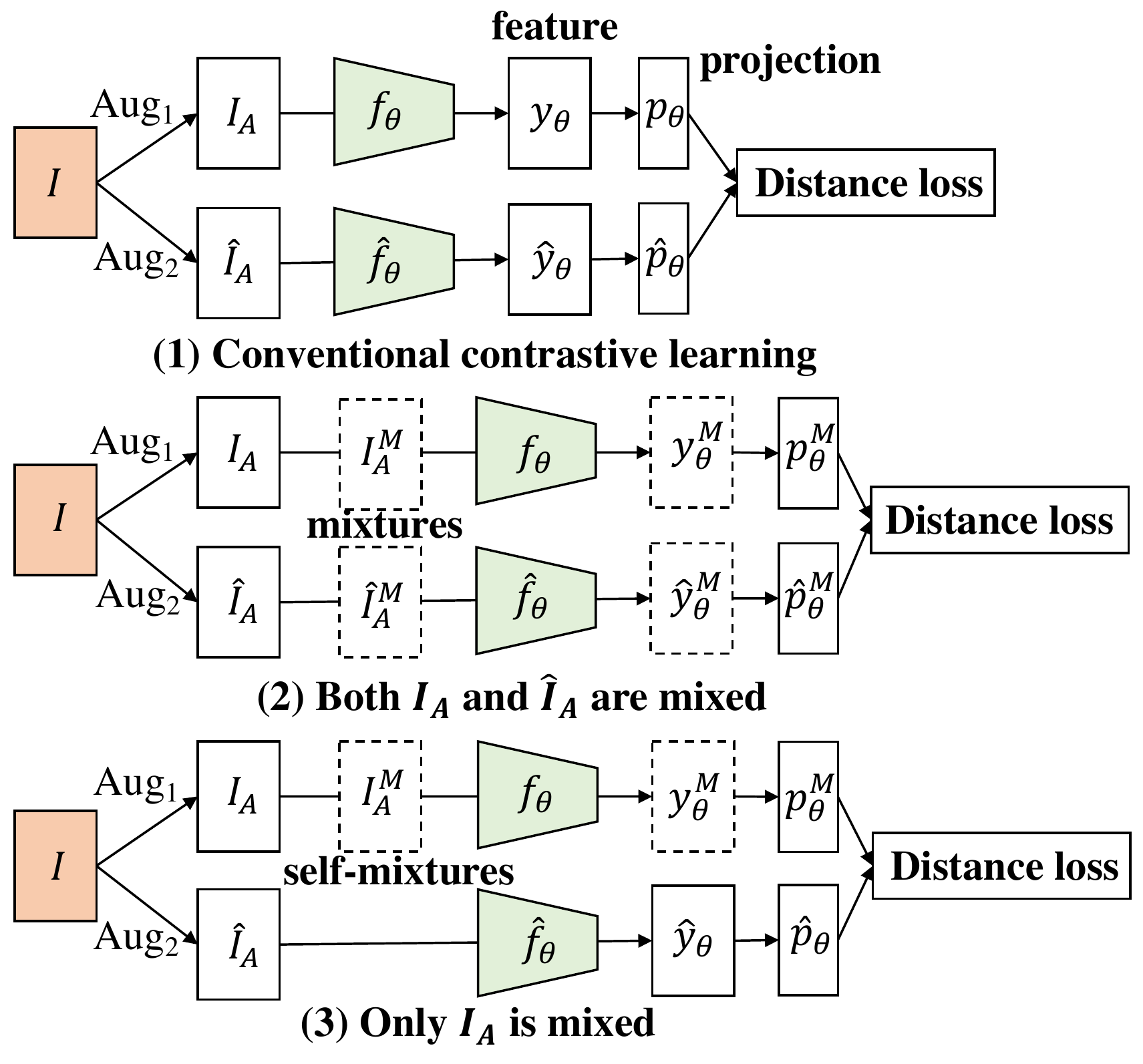}
  \vspace{-0.12in}
  \caption{Comparison of different paradigms of utilizing mixtures in unsupervised learning. (1) is the conventional instance classification based framework. (2) and (3) are the strategies of applying the proposed image mixtures. ``self-mixtures'' denotes that the images of mixture operation only happens in current batch samples. The dashed bounding box represents the mixed image and its representation.}
  \label{fig:paradigms}
  \vspace{-0.12in}
\end{figure}

\vspace{-0.05in}
The proposed mixtures follow the image transformations of input samples. We define $I^M_A$ and $\hat I^M_A$ as the mixed images which can be $\{I_{\bf{g\_m}}, I_{\bf{r\_m}\}}$ (global and region-level image mixtures respectively) according to the type of mixture operation we choose in the current training iteration. The mixed images are forwarded through the target network $f_\theta$, then a non-linear projection head $p_\theta$ is adopted to obtain the representations of the input sample for the unsupervised distance loss. Image mixture with relabeling can provide additional subtle information to force two branches unequally distant, instead of solely learning {\em positive} or {\em negative} pairs for the representations. In the following, we discuss two circumstances in such a framework, as shown in Fig.~\ref{fig:paradigms} (2) (3).

\noindent{\textbf{Both {\bf $I_A$} and $\hat I_A$ are mixed (Fig.~\ref{fig:paradigms} (2)).}} This solution is to mix both the two views of an input image. Thus, the similarity between mixtures will remain unchanged. While this  mixture strategy on two branches will suffer from undesirable equilibria as the mixture ratio of images is not used on loss space, namely, the additional information of mixture degree is not fully utilized. We found this strategy is effective on relatively small-scale datasets like CIFAR, STL-10, {\em etc.}, but is barely helpful on the large-scale ImageNet.

\noindent{\textbf{Only {$I_A$} is mixed (Fig.~\ref{fig:paradigms} (3)).}} This is the main strategy that we use in this work. Compared to the one above, this solution is more efficient since it only needs one additional forward pass. Also, reverse order outputs can be obtained by permutation from normal order outputs. From our experimental results it is also more effective for obtaining accuracy gain. 

\vspace{-0.04in}
\subsection{3.2. Image Mixture Strategies}

We introduce two widely-used mixture methods in supervised learning: \textbf{(i) Mixup}~\cite{zhang2018mixup} and \noindent{\textbf{(ii) Cutmix}}~\cite{yun2019cutmix}. Since they are designed for supervised learning with available ground-truth for calculating mixed labels, in this work, we focus on exploring the way to sample training data in a {\em mini}-batch and assign new softened distance loss formulations in the unsupervised learning frameworks.

\noindent{\textbf{Mixup}} can be written as: \vspace{-0.06in}
\begin{equation}\label{g_mix}
\begin{array}
{l}{I_{g\_m} \leftarrow \alpha I_{1}+(1-\alpha) I_{2}} 
\end{array}
\vspace{-0.04in}
\end{equation}
where $\{I_{1}, I_{2}\}$ denote the images that we want to mix. $I_{g\_m}$ is the output mixture, $\alpha \in[0,1]$ is the mixture coefficient.

\noindent{\textbf{Cutmix}} replaces within particular locations of a region: \vspace{-0.04in}
\begin{equation}\label{r_mix}
\begin{array}
{l}{I_{r\_m} \leftarrow {\bf M}_{b} \odot I_{1}+ ({\bf 1}-{\bf M}_{b}) \odot I_{2} }
\end{array}
\vspace{-0.04in}
\end{equation}
where ${\bf M}_b\in\{0,1\}^{I}$ denotes a binary mask as defined in~\cite{yun2019cutmix}. $\bf 1$ is a binary mask with all values equaling one. $\odot$ denotes element-wise multiplication.

Both Mixup and Cutmix can be regarded as the regularization techniques to prevent the models from overfitting and make the predictions less confident.

\noindent{\textbf{Dealing with Memory Banks (MB).}}
In this part, we describe different scenarios regarding how to design the framework using the proposed mixture training strategy if the base model contains a memory bank or not. Our goal is to enhance visual feature representations by leveraging additional mixture information and the different mixing ratios between two images in the unsupervised scheme. To this end, we propose a way to re-measure the distance of one pair samples for the MB-based or non-MB-based unsupervised frameworks.

\noindent{\textbf{(i) Without a memory bank.}} Under this circumstance, the unsupervised frameworks will use positive pairs only for training ({\em e.g.}, BYOL~\cite{grill2020bootstrap}) or contrastive-based pipelines ({\em e.g.}, SimCLR~\cite{chen2020simple}). Therefore, we only need to design the new distance of the positive pairs, as shown in Fig.~\ref{fig:distance_}. In our proposed self-mixture strategy, the new distance scale ${\mathcal{D}_{\operatorname{dis}}}$ of a positive pair will be:
\vspace{-0.02in}
\begin{equation}
\vspace{-0.02in}
\begin{array}{c}I_{A}^{M}=\lambda I_{1}+(1-\lambda) I_{2}, \\
{\mathcal{D}_{\operatorname{dis}}}\left(I_{A}^{M}, \hat{I}_{A}\right)=\left\{\begin{array}{ll}\lambda & \text { if }\ \  \hat{I}_{A}=\hat I_{1}, \\ 1-\lambda & \text { if } \ \ \hat{I}_{A}=\hat I_{2}.\end{array}\right.
\vspace{-0.08in}
\end{array}
\end{equation}
where $\hat I_1$, $\hat I_2$ are another views of $I_1$, $I_2$ from the same images. In traditional unsupervised scheme, they are a positive pair. $\lambda$ is the mix ratio controlled by the degree of mixture we use in the current iteration of training, when employing global mixture, $\lambda=\alpha$ as in Eq.~\ref{g_mix}, otherwise, $\lambda=\frac{{\bf M}_b}{\bf 1}$ as in Eq.~\ref{r_mix}.

\noindent{\textbf{(ii) With a memory bank.}} Using a memory bank with mixtures will solely affect the constitution of negative pairs as the distance/label between them is always ``zero'' in instance classification based contrastive learning. We keep the distance of negative pairs as original values, whatever they are the combination of one original and one mixed images. In particular, negative pairs (samples) can be {\em \{original, original\}}, {\em \{original, mixed\}}, {\em \{mixed, mixed\}} images. We found in experiments that maintaining one MB with the representations from original/unmixed images is enough to obtain good performance, we explain this through the enlarged training data space. However, this is inapplicable in the multi-scale training scheme, as we will discuss later in our Appendix. 

\begin{figure}[h]
  \centering
  \vspace{-0.13in}
  \includegraphics[width=0.47\textwidth]{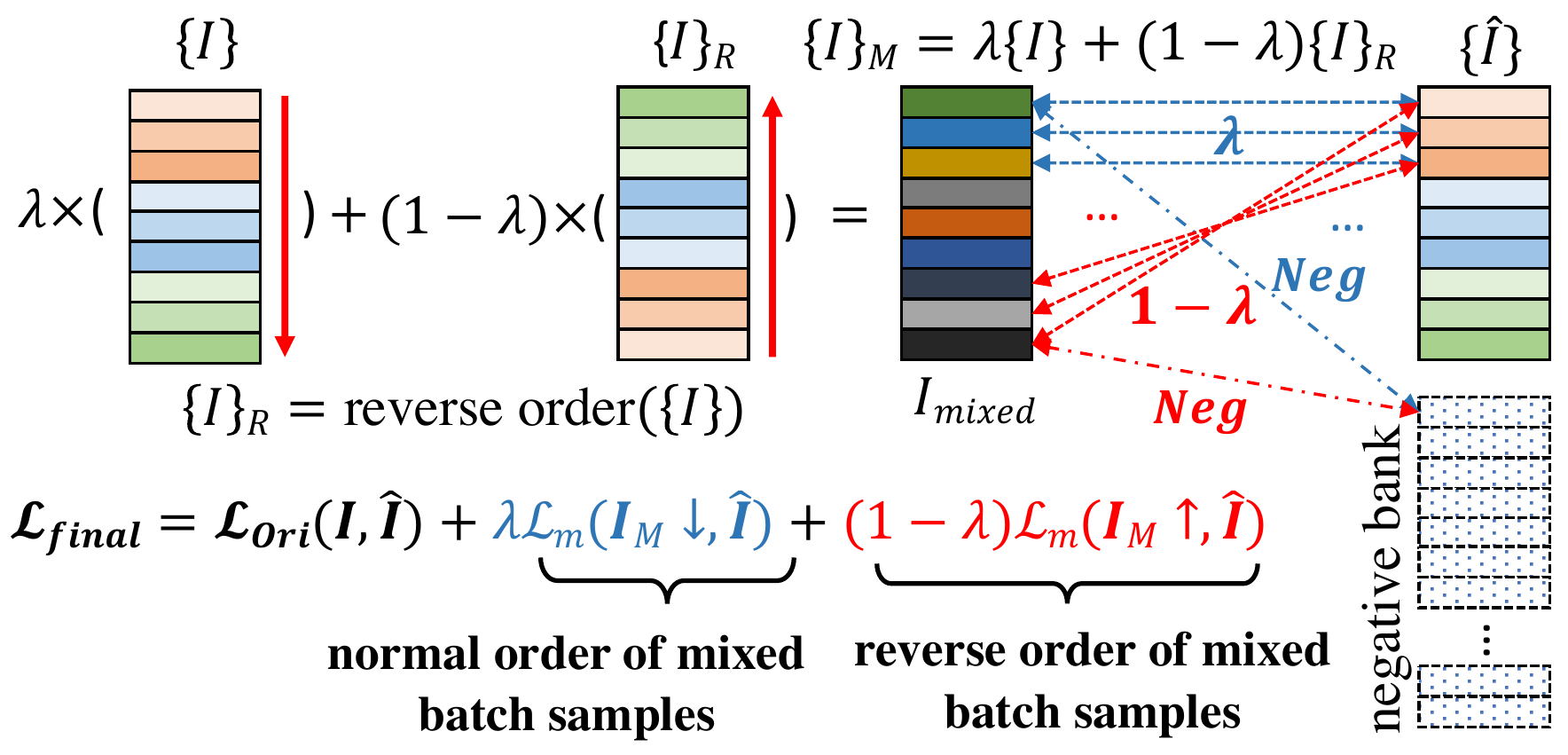}
  \vspace{-0.26in}
  \caption{Illustration of self-mixture within a {\em mini}-batch. In each iteration, we randomly choose one mixture operation for all the current samples with a pre-defined probability $P$, thus the formulation of $\lambda$ depends on the chosen mixture type.}
  \label{fig:self_mixture}
  \vspace{-0.12in}
\end{figure}

\vspace{-0.08in}
\subsection{3.3. Loss Functions} \label{loss_f}

\begin{figure}[t]
  \centering
  \includegraphics[width=0.24\textwidth]{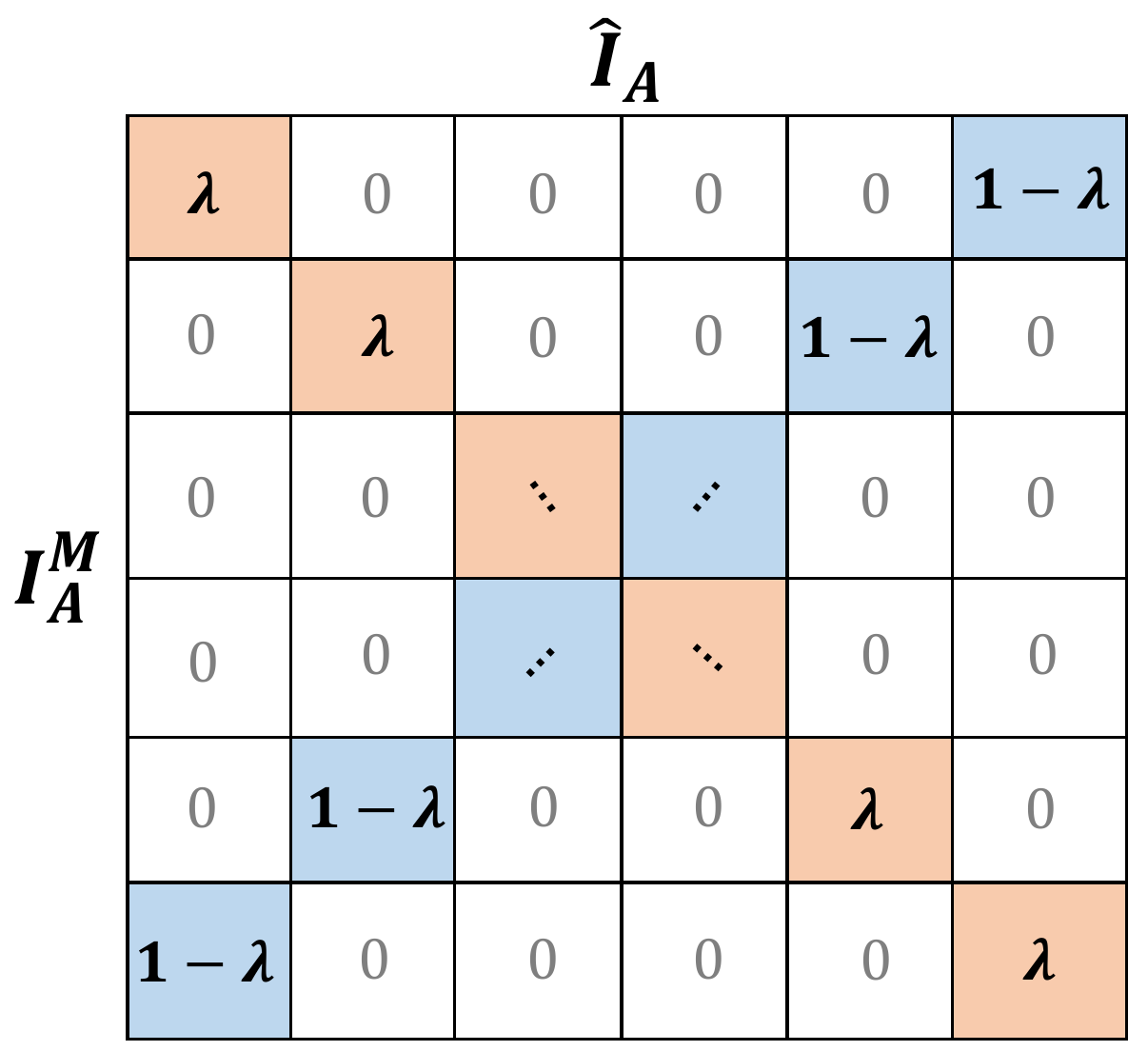}
  \vspace{-0.1in}
  \caption{The distance matrix of proposed mixture strategy between the mixed $I_A$ ({\em i.e.}, $I^M_A$) and $\hat I_A$ for calculating the softened distance loss. Here we take six images in the {\em mini}-batch as an example.}
  \label{fig:distance_}
  \vspace{-0.2in}
\end{figure}

\noindent{\textbf{Self-Mixtures within Per {\em Mini}-Batch of Training.}}
In our method, the mixing ratio of two images is the core gifted extra information that can be utilized in the unsupervised methods. Also, properly proposing a strategy to reflect the image mixture information in the loss space is crucial for leveraging image mixture in the self-supervised domain. Here we introduce the strategy of how to retain such information for loss calculation. We propose to directly mix the first image with the last one in each {\em mini}-batch of training, the second one is mixed with the penultimate, and so on. Our strategy is visualized in Fig.~\ref{fig:self_mixture} and Fig.~\ref{fig:distance_}, the advantages of such a strategy are: {\textbf{(i)}} Different from employing individual ratio for each image in one {\em mini}-batch, the proposed scheme can be realized through calculating the batch loss with a weighted coefficient, which is well-regulated, manageable, more efficient for implementing and can facilitate the design of label assignment in unsupervised frameworks. {\textbf{(ii)}} The proposed strategy will make the soft distances between the mixtures and original samples to be consistent across all pairs within a {\em mini}-batch. Hence, the calculation rule of loss function will be simplified and independent from the different frameworks that are employed, for instance, contrastive learning frameworks that use both positive and negative pairs or positive only, memory bank or without it, {\em etc.}, as scaling similarity distance is equivalent to weighting these loss values.

We now elaborate the loss functions. We compute an extra loss from a mixed pair of images. Given two mixed images $I^M_A$ and $\hat I^M_A$ from two
different augmentations of the same image (the case that both branches are mixed), we compute their loss together with the original one as the following:
\vspace{-0.1in}
\begin{equation}
\mathcal{L}_{both}=\mathcal{L}_{ori}(I_A, \hat I_A)+\underbrace{ \mathcal{L}_{m}(I^{M}_A, \hat I^M_A)}_{\text {extra term of mixtures}}
\vspace{-0.1in}
\end{equation}
where $\mathcal{L}_{ori}$ is the original loss function corresponding to the base method we use, like InfoNCE, $\ell_2$ distance, {\em etc.}, and $\mathcal{L}_{m}$ measures the fit between samples $I^M_A$ and $\hat I^M_A$.\\
Finally, we define the following sum of three loss terms from the original and mixed predictions as the ultimate objective:
\vspace{-0.1in}
\begin{equation}   
\mathcal{L}_{final}\!=\!\mathcal{L}_{ori}+\underbrace{\lambda \mathcal{L}_{m}\!(I_A^{M} (\downarrow), \hat I_A)}_{\text {normal order of mixtures}}\!+\!\underbrace{(1\!-\!\lambda) \mathcal{L}_{m}\!(I_A^{M} (\uparrow), \hat I_A)}_{\text {reverse order of mixtures}} 
\end{equation}
where the last two loss terms measure the fit between samples $I^M_A$ and $\hat I_A$, as detailed above. 
We take contrastive loss with InfoNCE as an example (notations refer to MoCo):
\vspace{-0.1in}
\begin{equation}   
\begin{aligned}
	\underbrace{ \mathcal{L}_{m}\!(I_A^{M} (\downarrow), \hat I_A)}_{\text {normal order of mixtures}}\!=\!-\log \frac{\exp \left(q_m \cdot k^{*} / \tau\right)}{\sum_{i=0}^{K} \exp \left(q_m \cdot k_{i} / \tau\right)};\\ \!\underbrace{ \mathcal{L}_{m}\!(I_A^{M} (\uparrow), \hat I_A)}_{\text {reverse order of mixtures}} = -\log \frac{\exp \left(q_{rm} \cdot k^{*} / \tau\right)}{\sum_{i=0}^{K} \exp \left(q_{rm} \cdot k_{i} / \tau\right)}
\end{aligned}
\vspace{-0.05in}
\end{equation}
where $q_m$, $q_{rm}$ are normal/reverse orders of mixed queries in a {\em mini}-batch, $k^{*}$ is the unmixed single key, $\tau$ is temperature.

\begin{figure*}[t]
  \centering
  \includegraphics[width=0.89\textwidth]{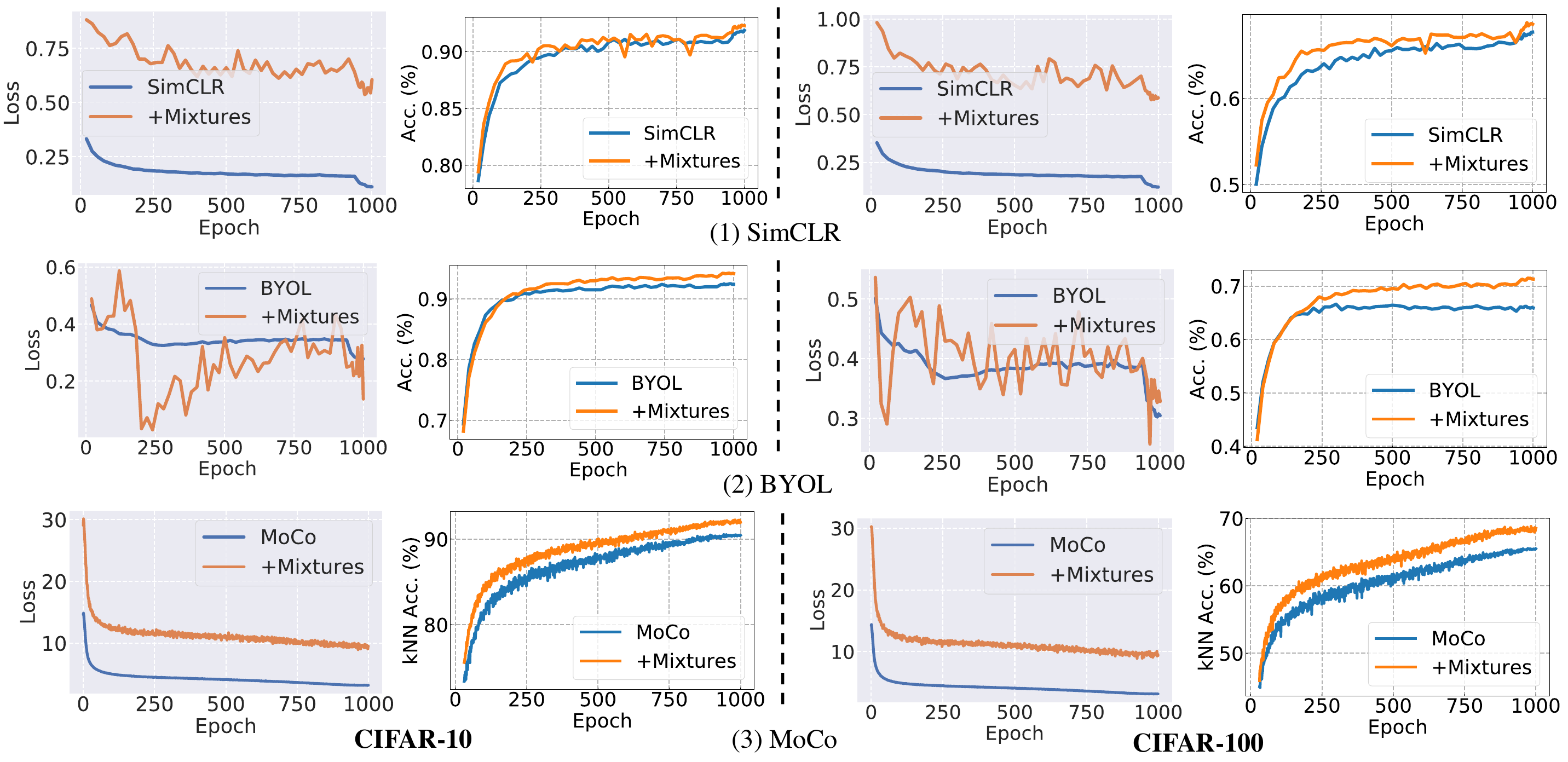}
  \vspace{-0.1in}
  \caption{Training loss and testing accuracy of SimCLR, BYOL and MoCo on CIFAR-10/100.}
  \label{fig:single_non_imagenet}
  \vspace{-0.15in}
\end{figure*}

\noindent{\textbf{Justifications from the Mutual Information (MI) Theory.}}
Suppose the latent representations of inputs are calculated as ${\bm z}_{{ori}}=f_{\theta_{1}}\left({I}_{{ori}}\right), {\bm z}_{{mix}}=f_{\theta_{2}}\left({I}_{{mix}}\right)$. According to~\cite{oord2018representation}, the mutual information ${\bm I}({\bm z}_{{ori}}, {\bm z}_{{mix}})$ of InfoNCE loss can be formulated as:
\vspace{-0.06in}
\begin{equation}
{\bm I}\left({\bm z}_{{ori}}, {\bm z}_{ {mix}}\right) \geq \log (\bm N)-\mathcal{L}_{\bm N}
\vspace{-0.04in}
\end{equation}
where $\bm N$ is the number of training samples (one positive and $\bm N$-1 negative samples). To maximize the lower bound of MI, one way is to minimize the InfoNCE objective $\mathcal L_{\bm N}$, while, ${\bm I}$ can also increase when $\bm N$ becomes larger which equivalently maximizes a lower bound on ${\bm I}({\bm z}_{{ori}}, {\bm z}_{ {mix}})$. Considering contrastive pairs without mixture, we build an $\binom{n}{1}$ relationships ($n$ is the number of images) in the dataset, only ($n$-$1$) images are negative pairs to the original one. After adding mixtures of two images, the MI we utilized is $\!\binom{n}{2}$ relationships. In general, using additional mixtures (equivalent to enlarge values of $\bm N$) does increase the tightness of mutual information $\bm I$.

\vspace{-0.03in}
\section{4. Experiments}
We demonstrate the effectiveness and superiority of our Un-Mix learned models with unsupervised pretraining on a variety of datasets. We first evaluate the representation ability in linear evaluation protocol. We then measure its transferability using object detection task on PASCAL VOC and COCO. 

\vspace{-0.06in}
\subsection{4.1. Datasets} \label{dataset}

\noindent{\textbf{CIFAR-10/100}~\cite{krizhevsky2009learning}.}
The two CIFAR datasets consist of tiny colored natural images with a size of 32$\times$32. CIFAR-10 and 100 are drawn from 10 and 100 classes, respectively. In each dataset, the train and test sets contain 50,000 and 10,000 images. 

\noindent{\textbf{STL-10}~\cite{coates2011analysis}.} STL-10 is inspired by CIFAR-10 with 10 classes, but each class has fewer labeled training examples (500 training images and 800 test images per class). The size of images is 96$\times$96. It also contains 100,000 unlabeled images for unsupervised learning.

\noindent{\textbf{Tiny ImageNet.}}
Tiny ImageNet is a lite version of ImageNet which contains 200 classes with images resized down to 64$\times$64. The train and test sets contain 100,000 and 10,000 images, respectively.

\noindent{\textbf{ImageNet}~\cite{deng2009imagenet}.}
The ImageNet, aka ILSVRC 2012 classification dataset~\cite{deng2009imagenet} consists of 1000 classes, with a number of 1.28 million training images and 50,000 validation images.

\begin{table*}[t]
\centering
\begin{minipage}{0.33\textwidth}
  \centering
  \includegraphics[width=0.9\textwidth]{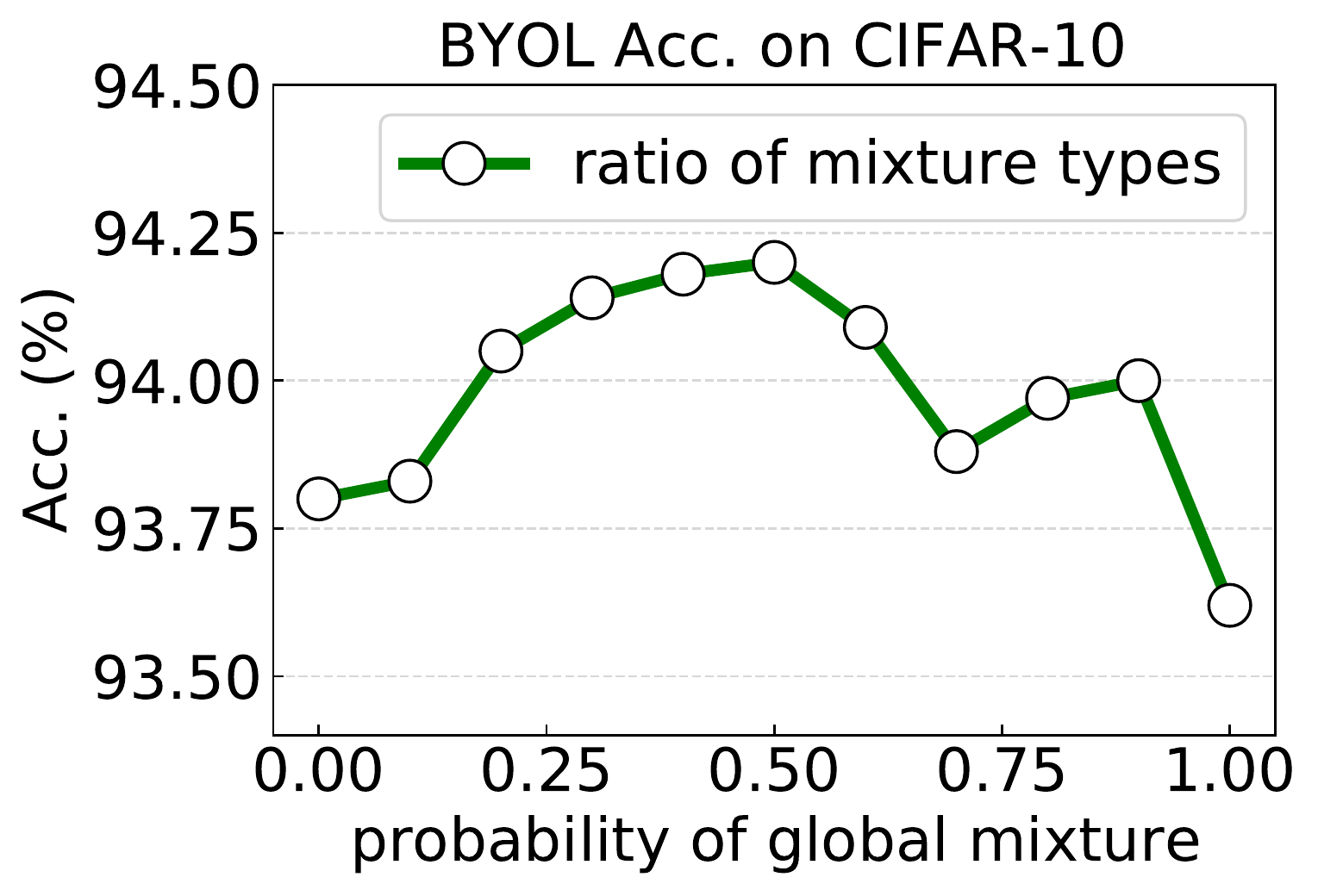}
  \vspace{-0.13in}
  \captionof{figure}{Acc. with various $P$.}
  \label{fig:tradeoff}
\end{minipage}
%
\begin{minipage}{0.28\textwidth}
    \vspace{-0.16in}
    \centering
    \small
    \setlength{\tabcolsep}{2pt}
\caption{Sensitivity for $\gamma$.}
 \vspace{-0.12in}
\label{tab:gamma}
{
\renewcommand{\arraystretch}{1.25}
\resizebox{0.85\textwidth}{!}{
\begin{tabular}{|c|c|c|c|}
\hline
CIFAR-10  & \multicolumn{3}{c|}{$\gamma$ in {\bf Beta} sampling} \\ \hline
          & 1.0     & 0.8    & 0.5    \\ \hline
Acc. (\%) & \bf 94.20   & 94.12  & 93.93  \\ \hline
\end{tabular}
}

 \vspace{-0.06in}
\caption{Sensitivity for $P$.}
\label{tab:tradeoff2}
\resizebox{0.85\textwidth}{!}{
\begin{tabular}{|c|c|c|c|}
\hline
ImageNet  & \multicolumn{3}{c|}{$P$ of global mixture}  \\ \hline
          & 1.0     & 0.5   & 0.0  \\ \hline
Acc. (\%) &  67.6  & 68.3  &  \bf 68.6  \\ \hline
\end{tabular}
}
}
\end{minipage} 
%
\begin{minipage}{0.38\textwidth}
\vspace{-0.2in}
\hspace{-0.10in}
  \includegraphics[width=1.05\textwidth]{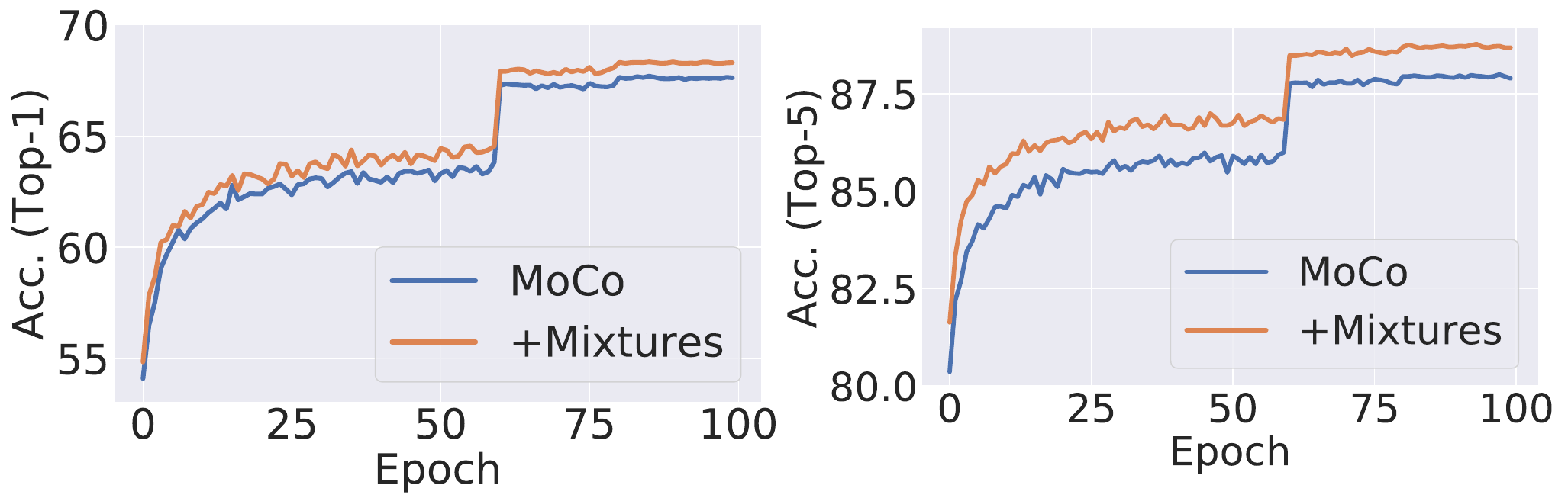}
  \vspace{-0.18in}
  \captionof{figure}{Linear classification accuracy of Top-1 (left) and Top-5 (right) with MoCo V2 and ours on ImageNet dataset.}
  \label{fig:mocov2}
  \vspace{-0.1in}
\end{minipage}

\vspace{-0.23in}
\end{table*}

\vspace{-0.06in}
\subsection{4.2. Baseline Approaches}

We perform our evaluation of image mixtures and label assignment strategy on the following five recently proposed unsupervised methods with state-of-the-art performance:

\noindent{\textbf{MoCo V1\&V2}~\cite{he2019momentum,MoCov2}.} MoCo is a contrastive learning method using momentum updating for unsupervised visual feature learning. MoCo V2 further improves momentum contrastive learning by adopting an MLP projection head and more/heavier data augmentation from the following SimCLR~\cite{chen2020simple}.

\noindent{\textbf{SimCLR}~\cite{chen2020simple}.} SimCLR is a simple framework for contrastive learning without requiring specialized architectures or a memory bank. It introduces a learnable nonlinear transformation that substantially improves the quality of the learned representations.

\noindent{\textbf{BYOL}~\cite{grill2020bootstrap}.} BYOL adopts online and target networks that learn from each other. It trains the online network to predict the target network representation of the same image under a different augmented view. At the same time, it updates the target network with a slow-moving average of the online network without the negative pairs.

\noindent{\textbf{SwAV}~\cite{caron2020unsupervised}.} SwAV is a clustering-based method for unsupervised learning. Unlike contrastive learning that compares features directly, it clusters the data while enforcing consistency between cluster assignments produced for different augmentations of the same image. 

\noindent{\textbf{Whitening}~\cite{ermolov2020whitening}.} Whitening is a loss function proposed for unsupervised representation learning which is based on the whitening of the latent space features. The whitening operation has a scattering effect to avoid degenerate solutions of collapsing to a simple status.\\
Our baseline approach implementations follow their official codebases which are all publicly available~\cite{facebookmoco,facebookmocodemo,se_supervised}.

\begin{figure}[t]
  \centering
  \includegraphics[width=0.46\textwidth]{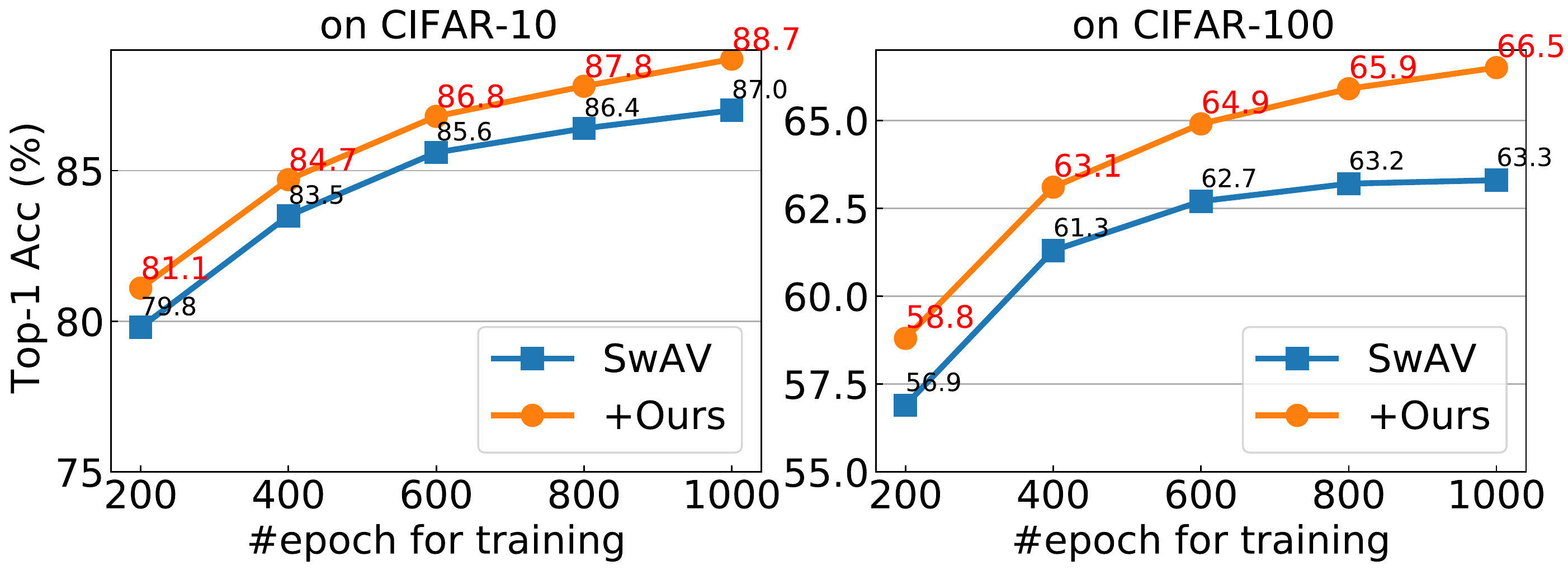}
  \vspace{-0.12in}
  \caption{Comparison under different training budgets.}
  \label{fig:swav}
  \vspace{-0.15in}
\end{figure}

\vspace{-0.08in}
\subsection{4.3. Implementation Details in Pre-training} \label{details}
The goal of our experiments is to demonstrate the effectiveness of our proposed image mixture and label assignment upon various unsupervised learning frameworks, isolating the effects of other settings, such as the architectural choices, data augmentations, hyper-parameters. As this, we use the same encoder ResNet-18 for all non-ImageNet experiments and ResNet-50 for ImageNet.
We use the same training settings, hyper-parameters, {\em etc.}, as our comparisons. Therefore, all gains in this paper are ``minima'', and further tuning the hyper-parameters in the baseline approaches to fit our mixture strategies might achieve more considerable improvement, while it is not the focus of this work.

\noindent{\textbf{Non-ImageNet Datasets.}}
Following~\cite{ermolov2020whitening}, on CIFAR-10 and CIFAR-100, we train for 1,000 epochs with learning rate 3$\times10^{-3}$; on Tiny ImageNet, 1,000 epochs with learning rate 2$\times10^{-3}$; on STL-10, 2,000 epochs with learning rate 2$\times10^{-3}$. We also apply warm-up for the first 500 iterations, and a 0.2 learning rate drop at 50 and 25 epochs before the end. 

\noindent{\textbf{Standard ImageNet.}}
Unless otherwise stated, all the hyperparameter configurations strictly follow the baseline MoCo V2 on ImageNet. For example, we use a {\em mini}-batch size of 256 with 8 NVIDIA V100 GPUs on ImageNet, considering our primary objective is to verify the effectiveness of proposed method instead of suppressing state-of-the-art results. 
For image mixtures and label assignment, we use $\gamma=1.0$ in beta sampling for all experiments, and $P=0.5$ for non-ImageNet and 0 for ImageNet based on our ablation study.

\subsection{4.4. Linear Classification}
Our linear classification experiments consist of two parts: {\bf (i)} ablation studies on small datasets including CIFAR-10, CIFAR-100, STL-10 and Tiny ImageNet with various base approaches to explore the optimal mixture hyperparameters and demonstrate the effectiveness of our strategy; {\bf (ii)} the final results on the standard ImageNet using MoCo V2.

\noindent{\textbf{Ablation Study.}}
We investigate the following aspects in our methods: {\bf (i)} the probability $P$ between global and region mixtures; {\bf (ii)} sensibility of $\gamma$ in beta distribution sampling. 

{\textbf{(1) Probability $P$ for choosing global or region-level mixtures in each iteration.}} The results are shown in Fig.~\ref{fig:tradeoff} (non-ImageNet) and Tab.~\ref{tab:tradeoff2} (ImageNet). They show that $P=0.5$ is optimal for small datasets and choosing region-level only ({\em i.e.}, $P=0$) is best for the large-scale ImageNet.

{\textbf{(2) Beta distribution hyperparameter $\gamma$.}} The combination ratio $\lambda$ between two sample points is sampled from the beta distribution $\textbf{Beta}(\gamma,\gamma)$. Our results on different $\gamma$s are presented in Tab.~\ref{tab:gamma}, $\gamma=1.0$ is the best and we use it for all our experiments, which means that $\lambda$ is sampled from a uniform distribution $[0, 1]$.

\noindent{\textbf{Results on CIFAR-10/100, STL-10 and Tiny ImageNet.}} Our results are shown in Tab.~\ref{table.non_imagenet} and Fig.~\ref{fig:single_non_imagenet}. All experiments are conducted on a single scale since the input sizes of these datasets are small, also for fair comparisons to baselines. Our method obtains consistency of $1\!\sim\!3$\% gains. In particular, our loss values are usually larger than the baselines (except BYOL, which is unstable since it has no negative pairs), but our accuracy is still superior. We also verify with different training budgets on SwAV, are shown in Fig.~\ref{fig:swav}. It can be seen our method still significantly benefits from longer training.

\noindent{\textbf{Results on ImageNet with MoCo V2.}} As shown in Tab.~\ref{tab:results_imgnet}, our method obtain 1.1\% improvement than baseline MoCo V2. Employing multi-scale training as in Appendix further boosts accuracy by 2.3\%. It is possible that tuning hyperparameters in MoCo V2, {\em e.g.}, temperature to fit our mixed training samples has potential to further improve performance.

\begin{table*}[t]
\begin{center}
\caption{Linear and 5-nearest neighbors classification results for different loss functions and datasets with a ResNet-18 backbone. Table is adapted from~\cite{ermolov2020whitening} and multi-scale training is not used for fair comparisons. Note that MoCo is trained with symmetric loss, 1000 epochs and evaluated with 200 in kNN monitor$^*$ following \href{https://colab.research.google.com/github/facebookresearch/moco/blob/colab-notebook/colab/moco_cifar10_demo.ipynb}{this}.}
\label{table.non_imagenet}
\vspace{-0.1in}
{\renewcommand{\arraystretch}{1.25}
\resizebox{0.999\textwidth}{!}{
\begin{tabular}{l|l l l l|l l l l|l l l l|l l l l}
\toprule
Method & \multicolumn{4}{c |}{CIFAR-10} & \multicolumn{4}{c |}{CIFAR-100} & \multicolumn{4}{c |}{STL-10} & \multicolumn{4}{c}{Tiny ImageNet} \\
 & linear & \bf ours & 5-nn & \bf ours & linear & \bf ours & 5-nn & \bf ours & linear & \bf ours & 5-nn & \bf ours & linear & \bf ours & 5-nn & \bf ours \\
\hline
SimCLR~\cite{chen2020simple} & 91.80 & \bf 92.35 & 88.42 & \bf 89.74 & 66.83 & \bf 68.83 & 56.56 & \bf 58.82 & 90.51 & \bf 90.86 & 85.68 & \bf 86.16 & 48.84 & \bf 49.58 & 32.86 & \bf 34.46 \\
BYOL~\cite{grill2020bootstrap}        & 91.73 & \bf 94.20 & 89.45 & \bf 93.03 & 66.60 & \bf 71.50 & 56.82 & \bf 63.83 & 91.99 & \bf 93.34 & 88.64 & \bf 90.46 & 51.00 & \bf 53.39 & 36.24 & \bf 39.27\\
Whitening (W = 2)~\cite{ermolov2020whitening}     & 91.55 & \bf 93.04 & 89.69 & \bf 91.33 & 66.10 & \bf 70.12& 56.69 & \bf 61.28 & 90.36 & \bf 92.21& 87.10 & \bf 88.88 & 48.20 & \bf 51.33 & 34.16 & \bf 36.78 \\
Whitening (W = 4)~\cite{ermolov2020whitening}     & 91.99 & \bf 93.18 & 89.87 & \bf 91.70 & 67.64 & \bf 69.70 & 56.45 & \bf 60.74 & 91.75 & \bf 91.96 & 88.59 & \bf 88.71  & 49.22 & \bf 50.67 & 35.44 & \bf 36.13 \\
MoCo (Sym. Loss)~\cite{he2019momentum}        & -- & --  & 90.49$^*$ & \bf 92.25$^*$ &  -- &  -- & 65.49$^*$ & \bf 68.83$^*$ & -- & -- & -- & -- & -- & -- & -- & --\\
\bottomrule
\end{tabular}}
}
\end{center}
\vspace{-0.25in}
\end{table*}

\begin{table}[t]
\newcommand*{\fct}[1]{\multicolumn{1}{>{\columncolor{white}\hspace*{-\tabcolsep}}c}{#1}}
\centering
\resizebox{0.48\textwidth}{!}{
\begin{tabular}{r|l|r|c|l}
\hline
\fontsize{8pt}{1em}\selectfont \bf Arch.
&\fontsize{8pt}{1em}\selectfont \bf Method 
& \multicolumn{1}{c|}{\fontsize{8pt}{1em}\selectfont \bf \#Params}
& \fontsize{8pt}{1em}\selectfont \bf Budget (\#ep) 
& \fontsize{8pt}{1em}\selectfont \bf Top-1 Acc. (\%) \\
\hline
R50 & MoCo  &  24\quad~ & 200 &  \quad 60.6 \\
R50 & CMC  &  24\quad~ & 200 &  \quad 66.2 \\
 R50 &  SimCLR  & 24\quad~ & 200 & \quad 66.6 \\
\rowcolor{mygraylite}  \cellcolor{white}  R50  &  MoCo V2 & 24\quad~ & 200 & \quad 67.5 \\
\rowcolor{mygraylite} \cellcolor{white} R50  & MoCo V2 + \bf Ours &24\quad~ & 200 & \quad \bf 68.6\color{darkergreen}$^{ \bf \uparrow1.1}$ \\
\rowcolor{mygraylite} \cellcolor{white}  R50  &MoCo V2 + \bf Ours$^{\dagger}$ &24\quad~ & 200 & \quad \bf69.8\color{darkergreen}$^{\bf\uparrow2.3}$ \\
 \hline
   R50  & PIRL & 24\quad~ & 800 & \quad 63.6 \\
  R50  & SimCLR & 24\quad~ & 1000 & \quad 69.3 \\
\rowcolor{mygray} \cellcolor{white}  R50  &MoCo V2 & 24\quad~ & 800 & \quad 71.1  \\
\rowcolor{mygray} \cellcolor{white} R50  &MoCo V2 + \bf Ours &24\quad~ & 800 & \quad 71.8\color{darkergreen}$^{\bf\uparrow0.7}$ \\
\hline
\end{tabular}
}
\vspace{-0.1in}
\caption{Comparison of linear classification on standard ImageNet. $^{\dagger}$denotes result using multi-scale training, more details can be referred to Appendix. Note that all the hyperparameters follow baseline MoCo V2 so  they might not be optimal on our mixture training, the gains are generally ``minima''.}.
\label{tab:results_imgnet}
\vspace{-0.2in}
\end{table}

\vspace{-0.05in}
\subsection{4.5. Downstream Tasks}
In this section, we evaluate the transferability of our learned representation on the object detection task. We use PASCAL VOC~\cite{everingham2010pascal} and COCO~\cite{lin2014microsoft} as our benchmarks and we strictly follow the same setups and hyperparameters of the prior works~\cite{he2019momentum,MoCov2} on the transfer learning stage. We use Faster R-CNN~\cite{ren2015faster} and Mask R-CNN~\cite{he2017mask} implemented in Detectron2~\cite{wu2019detectron2} with a ResNet-50~\cite{he2016deep} backbone. \\
\noindent{\textbf{PASCAL VOC.}}
We fine-tune our models on the split of \texttt{trainval07+12} and evaluate on the VOC \texttt{test2007} following~\cite{wu2018unsupervised,he2019momentum,misra2019self}. All models are fine-tuned for 24k iterations on VOC. 
It can be observed that significant improvements are consistently obtained by our proposed mixtures.

\noindent{\textbf{COCO.}}
We fine-tune on the \texttt{train2017} and evaluate on the \texttt{val2017} split. The total training budget is 180K iterations. The whole schedule follows the Detectron2 (\texttt{coco\_R\_50\_C4\_2$\times$}) default setting. Our results are shown in Tab.~\ref{tab:voc_coco} (b), 
it can be observed that our results are consistently better than the baseline by a significant margin.

\begin{table}[t]
\renewcommand{\randinit}{\tablestyle{1pt}{1} \begin{tabular}{z{21}y{26}} \multicolumn{2}{c}{Random init.} \end{tabular}}
\centering
\small
\subfloat[Faster R-CNN, \textbf{R50-C4} on {\bf PASCAL VOC}]{
\tablestyle{1pt}{1.1}
\resizebox{0.93\linewidth}{!}{
\begin{tabular}{x{80}|y{50}|y{50}|y{50}c}
\quad~Pre-train & 
\quad~AP$_{50}$&
\quad~AP & 
\quad~AP$_{75}$&\\ 
\shline
\randinit & \quad~{60.2} & \quad~{33.8} & \quad~33.1\\
\hline
Supervised IN-1M & \quad~{81.3} & \quad~{53.5} & \quad~58.8 \\
\rowcolor{mygraylite} MoCo V2 (200ep) & \quad~{82.4} & \quad~{57.0} & \quad~63.6 \\
\rowcolor{mygraylite} \bf Ours (200ep)  & \quad~\bf{83.0}\color{darkergreen}$^{\bf\uparrow0.6}$ & \quad~\bf{57.7}\color{darkergreen}$^{\bf\uparrow0.7}$ & \quad~\bf 64.3\color{darkergreen}$^{\bf\uparrow0.7}$  \\
\hline
\rowcolor{mygray} MoCo V2 (800ep) & \quad~{82.5} & \quad~{57.4} & \quad~64.0 \\
\rowcolor{mygray} \bf Ours (800ep)   & \quad~\bf{83.2}\color{darkergreen}$^{\bf\uparrow0.7}$ & \quad~\bf{58.1}\color{darkergreen}$^{\bf\uparrow0.7}$ & \quad~\bf 65.2\color{darkergreen}$^{\bf\uparrow1.2}$  \\
\hline
\bf Ours (200ep), MS & \quad~{83.2} & \quad~{57.8} & \quad~64.5 \\
\end{tabular}	}
} 

\vspace{-0.1in}

\subfloat[Mask R-CNN, \textbf{R50-C4} {\bf 2$\times$} on {\bf COCO}]{
\tablestyle{1pt}{1.0}
\resizebox{0.93\linewidth}{!}{
\begin{tabular}{x{80}|y{50}|y{50}|y{50}c}
Pre-train & 
\quad~AP&
\quad~AP$_{50}$&
\quad~AP$_{75}$& \\ 
\shline
\randinit & \quad~{35.6} & \quad~{54.6} & \quad~{38.2} & \\
\hline
Supervised IN-1M & \quad~{40.0} & \quad~{59.9} & \quad~43.1 \\
\rowcolor{mygraylite}MoCo V2 (200ep) & \quad~{40.9} & \quad~{60.7} & \quad~44.4 \\
\rowcolor{mygraylite}\bf Ours (200ep) 
& \quad~\bf{41.2}\color{darkergreen}$^{\bf\uparrow0.3}$ & \quad~\bf{60.9}\color{darkergreen}$^{\bf\uparrow0.2}$ & \quad~\bf 44.7\color{darkergreen}$^{\bf\uparrow0.3}$  \\
\end{tabular}	}
} 

\vspace{-0.12in}
\caption{Object detection results fine-tuned on PASCAL VOC (a) and COCO (b) datasets. 
Models are fine-tuned with the same number of iterations as the baseline, {\em e.g.,} 24k on VOC. 
On the VOC dataset, we run three trials and report the means.
}
\label{tab:voc_coco}
\vspace{-0.12in}
\end{table}

\vspace{-0.08in}
\subsection{4.6. Visualization and Analysis} \label{analysis}

\begin{figure}[t]
  \centering
  \includegraphics[width=0.48\textwidth]{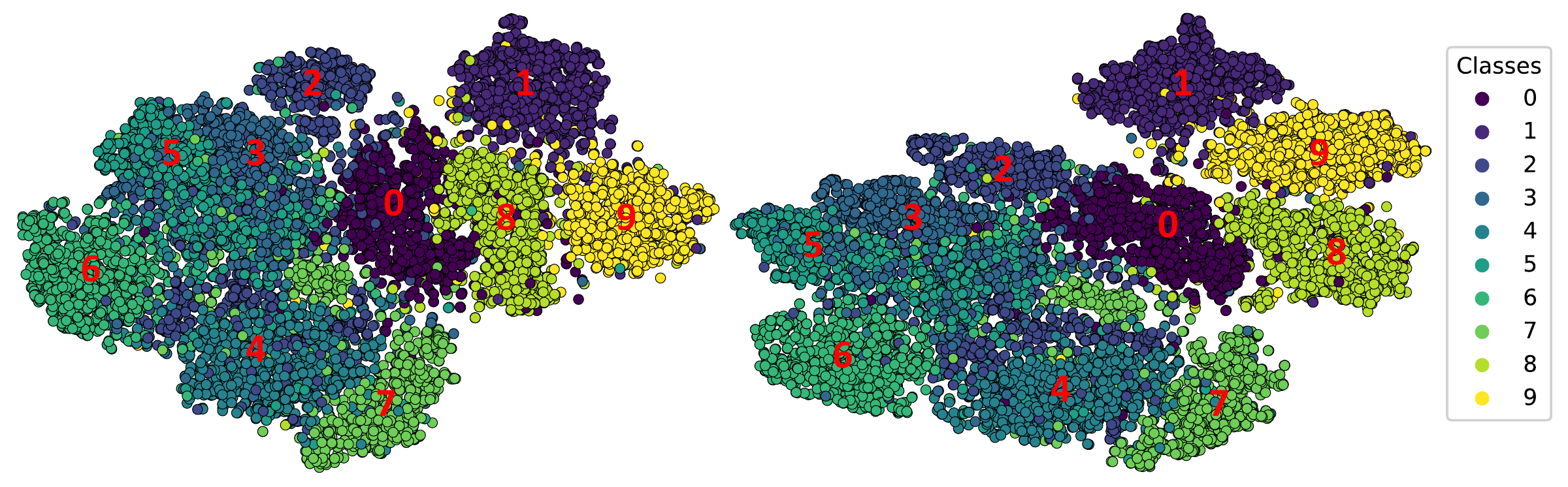}  
  \vspace{-0.2in}
  \caption{Visualizations of feature embeddings on CIFAR-10.}
  \label{fig:bem}
  \vspace{-0.13in}
\end{figure}

\noindent{\textbf{Learned representations.}} 
To further explore what our model indeed learned, we visualize the embedded features in Fig.~\ref{fig:bem} from baseline MoCo (left) and our mixture model (right) using t-SNE with the last conv-layer features (128-dimension) from ResNet-18. Our model has more separate embedding clusters, especially on classes 9, 8 and 1. We also visualize the histogram of weights in particular convolutional layers, as shown in our Appendix with discussions. 

\noindent{\textbf{Limitation.}} The only limitation we observed in our method is that it will take one additional forward pass for the mixed images. Since in our strategy, calculating the normal and reverse order of images' representations can share the same forwarding operation and there is no extra back-propagation, so the total extra cost will be less than one-third. We emphasize that the information gain of our method is from the mixed representations and mixture ratios, rather than the ``longer'' training. The mechanism of our method is different from longer training and cannot be replaced by it.

 \vspace{-0.04in}
\section{5. Conclusion}

We have investigated the feasibility of mixture operations in an unsupervised scheme, and proposed the strategy of image mixtures and corresponding label re-assignment for flattening inputs and predictions in various architectures of unsupervised frameworks. 
Through extensive experiments on SimCLR, BYOL, MoCo V1\&V2, {\em etc.}, and downstream tasks like object detection, we have shown that neural networks trained with our newly constructed input space have better representation capability in terms of generalization and transferability, as well as better robustness for different pretext tasks or frameworks (contrastive or non-contrastive learning, with or without memory banks, multi-scale training, {\em etc.}). Considering its simplicity to implement and it only incurs rational extra cost, we hope the proposed method can be a useful technique for the unsupervised learning problem.

\section*{Acknowledgement}

We thank all reviewers for their constructive and helpful comments in reviewing our paper. This manuscript has been revised significantly over its previous version.

{\small
\bibliography{aaai22}
}

\clearpage
\newpage

\appendix

\section*{\Large{\noindent{\textbf{Appendix}}}}
\vspace{1ex}

In this appendix, we provide details omitted in the main text, including:

• Section A ``Comparison of Different Loss Functions''. (Sec. ``Image Mixture Strategies'' of the main paper.)

• Section B ``Implementation Details'': Implementation details in unsupervised pre-training and linear evaluation on non-ImageNet and ImageNet. (Sec. ``Implementation Details in Pre-training'' and Sec. ``Linear Classification'' of the main paper.)

• Section C ``Multi-scale Training'': Discussions of {\em multi-scale training} using our mixture learning strategy. (Sec. ``Experiments'' of the main paper.)

• Section D ``Comparison of Different Mixture Strategies'': Comparison of {\em both branch mixtures} and {\em single branch mixture}. (Sec. ``Paradigms of Mixtures'' of the main paper.)

• Section E ``Pseudocode'': A PyTorch-like code for our mixture approach. (Sec. ``Introduction'' of the main paper.)

• Section F ``Visualization'': Visualization of entire layers with a ResNet-50 on ImageNet dataset. (Sec. ``Visualization and Analysis'' of the main paper.)

\begin{figure*}[t]
  \centering
  \includegraphics[width=0.71\textwidth]{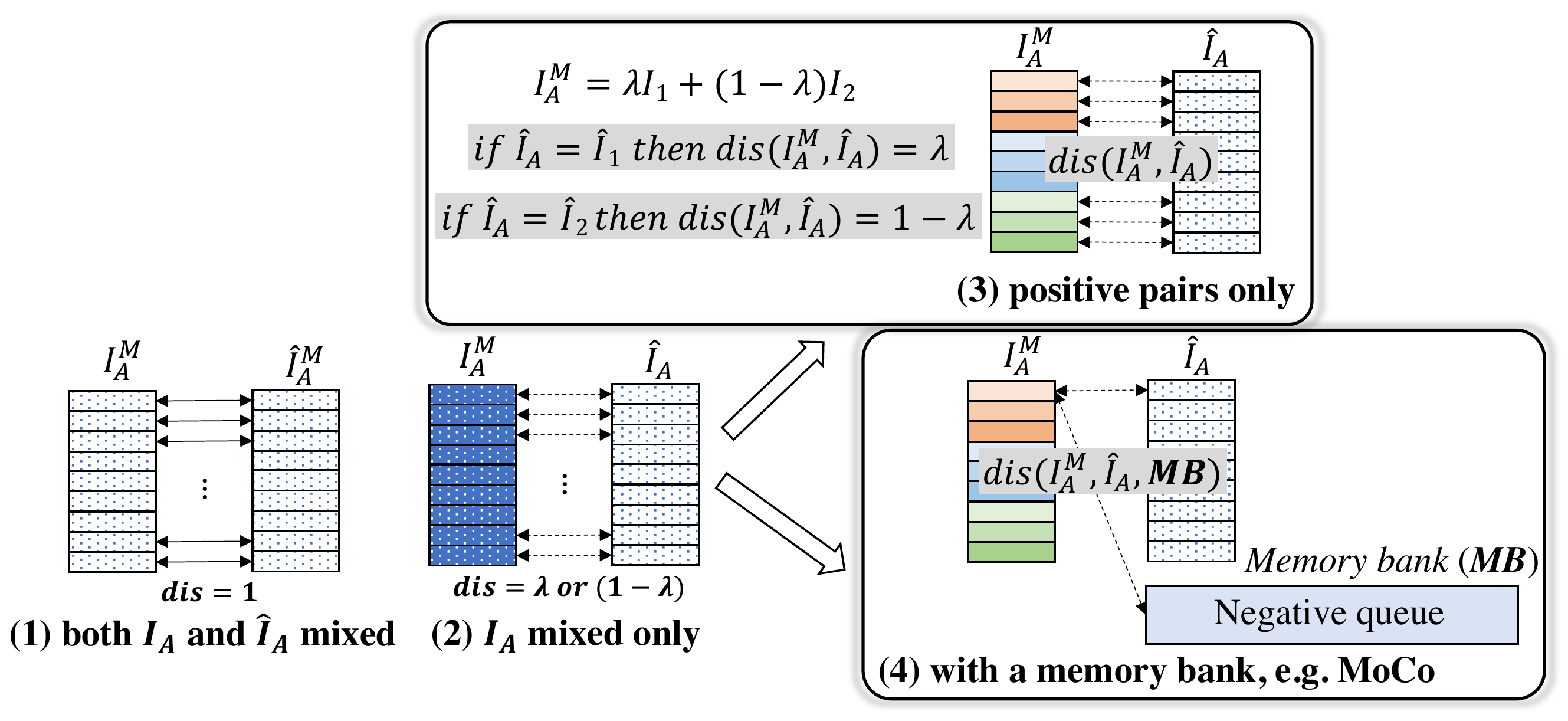}
  \vspace{-0.08in}
  \caption{Illustration of our mixture strategies (both branches are mixed or only $I_A$ is mixed). We consider the unsupervised frameworks (i) contain a memory bank or not; (ii) train with positive pairs only or also involve negative pairs.}
  \label{fig:motivation}
\end{figure*}

\subsection{A. Comparison of Different Loss Functions}
\label{tab:results_imgnet}
The accuracies when applying different loss functions (using MoCo V2 default loss and our Un-Mix loss on ImageNet) are shown in Tab.~\ref{tab:results_imgnet}. We can observe that the original contrastive loss (MoCo V2) can obtain 67.5\% on ImageNet. Using mixed image pairs solely without the unmixed image pairs ({\em i.e.,} original images) for training can also achieve a decent and similar accuracy (66.8\%), this proves that the mixed image pairs is already an adequate and strong design for learning good representations in the siamese-like frameworks. Applying both of them can further improve the performance to 68.6\% with 1.1\% boost over the baseline.

\begin{table}[h]
\newcommand*{\fct}[1]{\multicolumn{1}{>{\columncolor{white}\hspace*{-\tabcolsep}}c}{#1}}
\centering
\resizebox{0.48\textwidth}{!}{
\begin{tabular}{r|l|r|c|l}
\hline
\fontsize{8pt}{1em}\selectfont \bf Arch.
&\fontsize{8pt}{1em}\selectfont \bf Method 
& \multicolumn{1}{c|}{\fontsize{8pt}{1em}\selectfont \bf \#Params}
& \fontsize{8pt}{1em}\selectfont \bf Budget (\#ep) 
& \fontsize{8pt}{1em}\selectfont \bf Top-1 Acc. (\%) \\
\hline
R50 & $\lambda\mathcal{L}_m(\downarrow)$+$(1-\lambda)\mathcal{L}_m(\uparrow)$  &  24\quad~ & 200 &  \quad 66.8 \\
R50 & $\mathcal{L}_{ori}$  &  24\quad~ & 200 &  \quad 67.5 \\

\rowcolor{mygraylite} \cellcolor{white} R50  & $\mathcal{L}_{ori}$ + $\lambda\mathcal{L}_m(\downarrow)$+$(1-\lambda)\mathcal{L}_m(\uparrow)$&24\quad~ & 200 & \quad \bf 68.6\color{darkergreen}$^{ \bf \uparrow1.1}$ \\
\hline
\end{tabular}
}
\vspace{-.1in}
\caption{Comparison of linear classification using different loss functions on standard ImageNet.}.
\label{tab:results_imgnet}
\vspace{-0.2in}
\end{table}

\begin{algorithm}[h] 
\caption{PyTorch-like Code for Un-Mix Strategy.}
\label{alg:code}
\definecolor{codeblue}{rgb}{0.25,0.5,0.5}
\lstset{
  backgroundcolor=\color{white},
  basicstyle=\fontsize{7.2pt}{7.2pt}\ttfamily\selectfont,
  columns=fullflexible,
  breaklines=true,
  captionpos=b,
  commentstyle=\fontsize{7.2pt}{7.2pt}\color{codeblue},
  keywordstyle=\fontsize{7.2pt}{7.2pt},
  moredelim=**[is][\color{red}]{@}{@},
}
\lstdefinestyle{base}{
  moredelim=**[is][\color{red}]{@}{@},
}
\begin{lstlisting}[language=python,,style=base]
# P: probability of global or local level mixtures
# beta: hyperparameter for Beta distribution
# lam: mixture ratio in global-level mixture or bounding box location in region-level mixture

args.beta = 1.0
for x in loader:  # load a minibatch x with N samples
    # Probability of choosing global or local level mixtures
    prob = np.random.rand(1) 
    lam = np.random.beta(args.beta, args.beta)
    images_reverse = torch.flip(x[0], (0,))
    if prob < args.P:
        # global-level mixture
        mixed_images = lam * x[0] + (1 - lam) * images_reverse
        mixed_images_flip = torch.flip(mixed_images, (0,))
    else:
        # region-level mixture
        mixed_images = x[0].clone()
        bbx1, bby1, bbx2, bby2 = utils.rand_bbox(x[0].size(), lam)
        mixed_images[:, :, bbx1:bbx2, bby1:bby2] = images_reverse[:, :, bbx1:bbx2, bby1:bby2]
        mixed_images_flip = torch.flip(mixed_images, (0,))
        lam = 1 - ((bbx2 - bbx1) * (bby2 - bby1) / (x[0].size()[-1] * x[0].size()[-2]))

    # original loss term
    loss_ori = model(x) 
    # loss for the normal order of mixtures
    loss_m1 = model([x[1], mixed_images]) 
    # loss for the reverse order of mixtures
    loss_m2 = model([x[1], mixed_images_flip]) 
    # final loss function
    loss = loss_ori + lam*loss_m1 + (1-lam)*loss_m2
    
    # update gradients
    optimizer.zero_grad()
    loss.backward()
    optimizer.step()
    ...
\end{lstlisting}
\end{algorithm}

\subsection{B. Implementation Details in Un-/self-supervised Pre-training and Linear Evaluation} \label{details}

Following settings in~\cite{ermolov2020whitening}, on CIFAR-10 and CIFAR-100, we train for 1,000 epochs with learning rate 3$\times10^{-3}$; on Tiny ImageNet, 1,000 epochs with learning rate 2$\times10^{-3}$; on STL-10, 2,000 epochs with learning rate 2$\times10^{-3}$. We also apply warm-up for the first 500 iterations, and a 0.2 learning rate decay at 50 and 25 epochs before the end of training. Adam is used as the optimizer in all of the non-ImageNet experiments. The weight decay is used as $10^{-6}$. The batch size is set to $K\!=\!512$ samples. The dimension of the hidden layer of the projection head is 1024. Moreover, we use an embedding size of 64 for CIFAR-10 and CIFAR-100, and 128 for STL-10 and Tiny ImageNet.

{\bf Evaluation Protocol on Non-ImageNet.} For linear evaluation on non-ImageNet datasets, we follow the baseline codebase configurations~\cite{ermolov2020whitening} with 500 epochs using the Adam optimizer without any data augmentation. The learning rate is exponentially decayed from 10$^{-2}$ to $10^{-6}$. The weight decay is $5\times10^{-6}$.

\begin{table*}[t]
\begin{center}
\caption{Classification accuracy of Top-1 on ResNet-18 using a linear classifier and a 5-nearest neighbors classifier with different loss functions and datasets. Ours$_1$ and ours$_2$ denote the results of {\em both branch mixtures} and {\em single branch mixture}, respectively. MoCo is trained with symmetric loss, 1000 epochs and evaluated with 200 in kNN monitor$^*$ following \url{https://colab.research.google.com/github/facebookresearch/moco/blob/colab-notebook/colab/moco_cifar10_demo.ipynb}.}
\label{table.sota_more}
\vspace{-0.08in}
{\renewcommand{\arraystretch}{1.5}
\resizebox{1.0\textwidth}{!}{
\huge
\begin{tabular}{l|l l l | l l l|l l l | l l l|l l l | l l l}
\toprule
Method & \multicolumn{6}{c |}{CIFAR-10} & \multicolumn{6}{c |}{CIFAR-100} & \multicolumn{6}{c }{STL-10} \\
 & linear & \bf ours$_1$  & \bf ours$_2$ & 5-nn & \bf ours$_1$  & \bf ours$_2$ & linear & \bf ours$_1$  & \bf ours$_2$ & 5-nn & \bf ours$_1$  & \bf ours$_2$ & linear & \bf ours$_1$  & \bf ours$_2$ & 5-nn & \bf ours$_1$  & \bf ours$_2$ \\
\hline
SimCLR~\cite{chen2020simple}      & 91.80 & \bf 93.11 & 92.35 & 88.42 & \bf 90.06 & 89.74 & 66.83 & \bf 69.47 & 68.83 & 56.56 & \bf 59.34 & 58.82 & 90.51 & \bf 91.16 & 90.86 & 85.68 & \bf 87.29& 86.16  \\
BYOL~\cite{grill2020bootstrap}        & 91.73 & 93.37 & \bf 94.20 & 89.45 & 91.86 &\bf 93.03 & 66.60 & 68.75 & \bf 71.50 & 56.82 & 61.21 & \bf 63.83 & 91.99 & 90.53 & \bf 93.34 & 88.64 & 87.68 & \bf 90.46  \\
W-MSE 2~\cite{ermolov2020whitening}     & 91.55 & 92.77 & \bf 93.04 & 89.69 & 91.06 & \bf 91.33 & 66.10 & 69.44 & \bf 70.12 & 56.69 & 59.16 & \bf 61.28 & 90.36 & 89.28 & \bf 92.21 & 87.10 & 85.59 & \bf 88.88  \\
W-MSE 4~\cite{ermolov2020whitening}     & 91.99 & 91.98 & \bf 93.18 & 89.87 & 89.85 & \bf 91.70 & 67.64 & 67.63 & \bf 69.70 & 56.45 & 57.46 & \bf 60.74 & 91.75 & 90.95 & \bf 91.96 & 88.59 & 87.86 & \bf 88.71  \\
MoCo (Symmetric Loss)~\cite{he2019momentum}        & -- & -- & -- & 90.49$^*$ & 90.35$^*$ & \bf 92.25$^*$ &  -- &  -- & -- & 65.49$^*$ & 66.01$^*$ & \bf 68.83$^*$ & -- & -- & -- & -- & -- & -- \\
\bottomrule
\end{tabular}}
}
\end{center}
\vspace{-0.1in}
\end{table*}

{\bf Evaluation Protocol on ImageNet.} For linear evaluation on ImageNet dataset, we follow the baseline approach~\cite{he2019momentum,MoCov2} and use an initial learning rate of 30 and weight decay of 0. We train with 100 epochs and the learning rate is multiplied by 0.1 at 60 and 80 epochs.

\subsection{C. Enabling Multi-scale Training}
\begin{figure}[t]
\vspace{0.02in}
  \centering
  \includegraphics[width=0.48\textwidth]{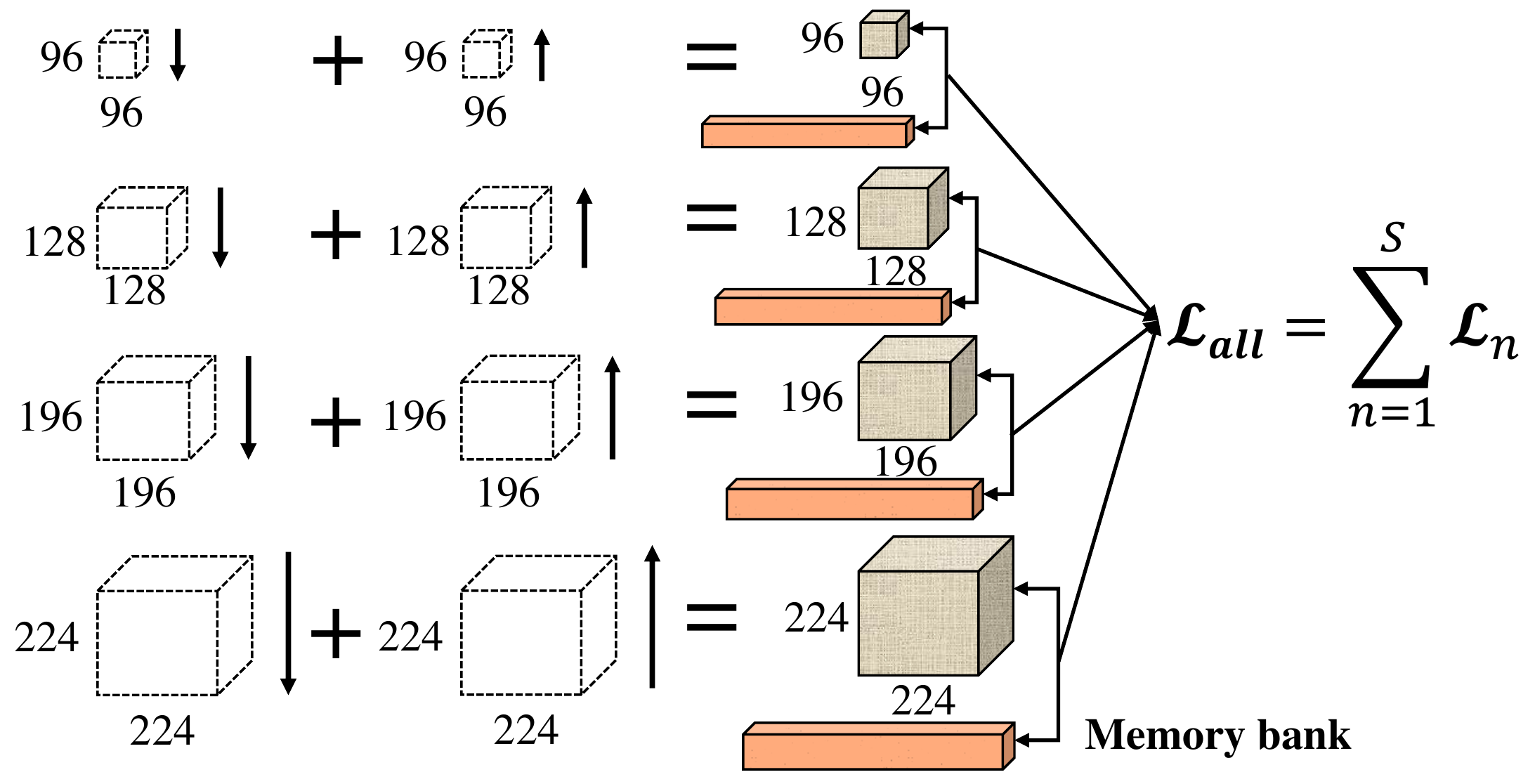}
  \vspace{-0.22in}
  \caption{Illustration of the proposed multi-scale training strategy. We aggregate losses from all of the different scales of inputs.}
  \label{fig:distance__}
  \vspace{-0.1in}
\end{figure}

Inspired by the prior study~\cite{caron2020unsupervised} that comparing random crops of an image is crucial for the networks to capture information of patterns from scenes or objects, we explore the possibility of employing multi-scale training in our mixture framework. Different from~\cite{caron2020unsupervised}, we discuss the circumstance of multi-scale training where the framework contains a memory bank. In such a framework, the input sizes of multi-scale training are different for a particular network, but after flowing forward through an MLP projection head, the latent features will be in the same dimension across different input sizes of branches, as shown in Fig.~\ref{fig:distance__}. As a memory bank is used to increase the number of negative pairs, so that for simplicity, we can share a single memory bank for all the various scales of inputs to generate negative pairs. Unfortunately, we found this sharing strategy leads to poor performance on representation ability, (this phenomenon is incongruous that we share the memory bank for original and mixed images in the single-scale training scenario of the main text). We conjecture that, in the multi-scale situation, such negative pairs are far away from the decision boundary so they are actually the ``easy'' negative samples. Adopting individual memory banks for each input size of samples is proven a better choice in our experiments. Note that this is a simple extension on Un-Mix and somewhat costly, we barely use it in the experiment whereas we will specify when we employed. As shown in Fig.~\ref{fig:distance__}, the final loss is aggregated from all the scales of training samples.

\begin{figure*}[t]
	\centering
	\includegraphics[width=0.99\textwidth]{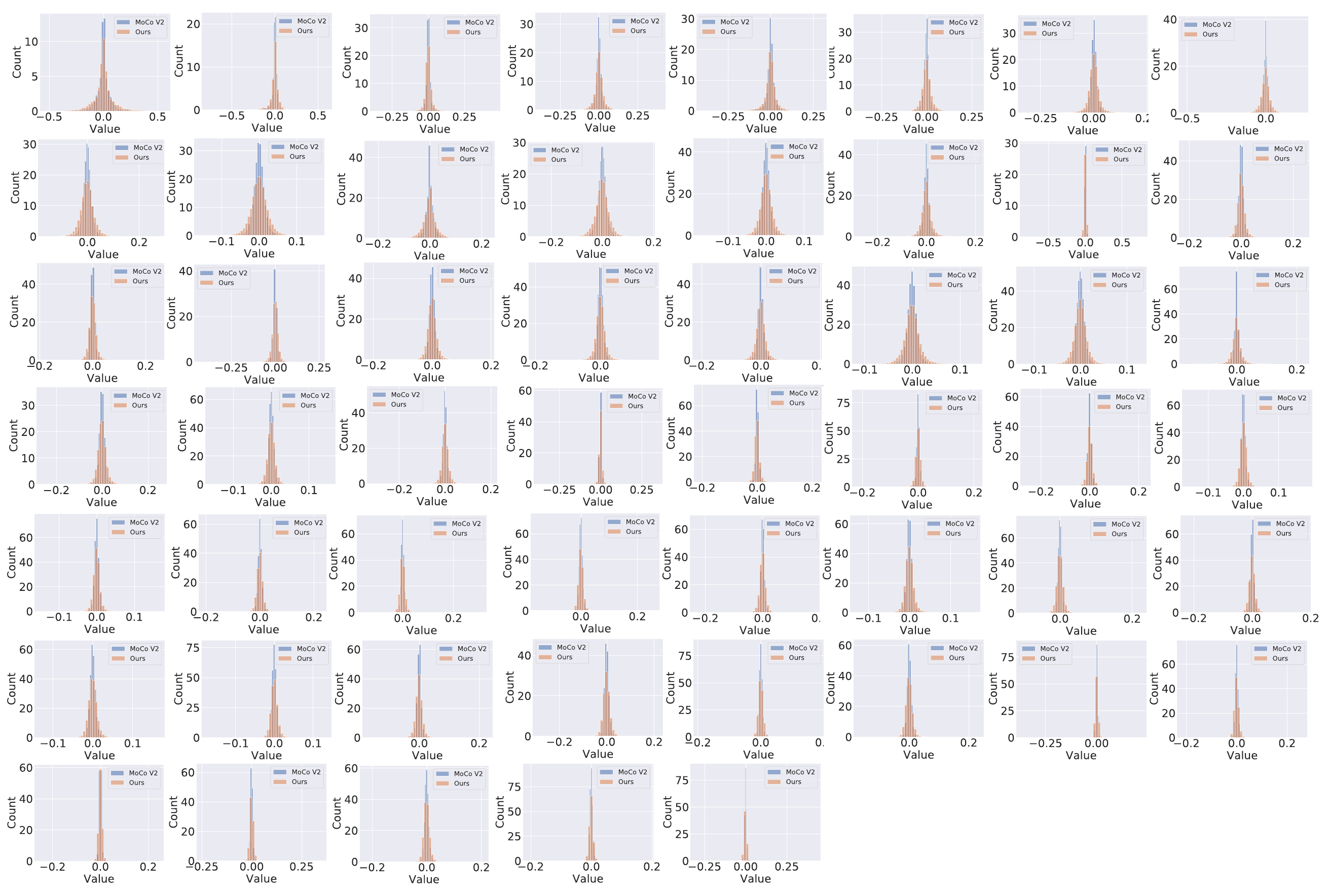}
	\vspace{-0.05in}
	\caption{Weight distributions from all the layers of a ResNet-50 architecture on ImageNet with baseline MoCo V2 and our Un-Mix trained model.}
	\label{fig:vis_weight}
\end{figure*}

\subsection{D. Comparison of {\em Both Branch Mixtures} and {\em Single Branch Mixture}}\label{lost_eff}

The comparisons between {\em both branch mixtures} and {\em single branch mixture} are provided in Fig.~\ref{fig:motivation} and Tab.~\ref{table.sota_more}, ``ours$_1$'' and ``ours$_2$'' denote the results of these two paradigms of mixtures, respectively. From this table, we have several interesting observations: We found that the strategy of {\em both branch mixtures}, i.e., ``ours$_1$'' performs better than ``ours$_2$'' on SimCLR~\cite{chen2020simple} across all datasets, but is inferior with other three baseline approaches. We conjecture the reason is that SimCLR leverages negative pairs and our mixtures can increase the capacity/complexity of composition on the negative pair set, which is beneficial for the unsupervised representation learning. However, the other three baseline approaches do not use negative pairs in their frameworks, so the advantage of enlarged capacity cannot be utilized by these methods. Also, ``ours$_1$'' has no additional mixture ratio information if comparing to ``ours$_2$'', so basically, ``ours$_2$'' is a superior design. Nevertheless, in most cases, both of these two paradigms are still better than the baseline results.

\subsection{E. A PyTorch-like~\cite{paszke2019pytorch} Code for Our Mixture Strategy} \label{code}

The pseudocode of our mixture method is shown in Algorithm~\ref{alg:code}. Note that this is a simple demonstration for the circumstance that the memory banks are not existing in the frameworks, such as {\em SimCLR}~\cite{chen2020simple}, {\em BYOL}~\cite{grill2020bootstrap} and {\em Whitening}~\cite{ermolov2020whitening} approaches. If employing a memory bank in the approach, it is required to deal with the memory bank mechanism so that it needs a few additional non-core lines of code, but the central implementation is entirely the same as we presented in Algorithm~\ref{alg:code}.

\subsection{F. Visualizations of All Convolutional Layers} \label{vis}

As shown in Fig.~\ref{fig:vis_weight}, we visualize the weight distributions from all convolutional layers of a ResNet-50 on ImageNet with the baseline MoCo V2 model and our mixture trained model. It can be observed that our weight distributions usually have a wider scope of values, while fewer elements are close to zero. This phenomenon reflects the potential larger capacity of our network to some extent since weights in our model have more status to be in. 

\end{document}